\documentclass[letterpaper, 10 pt, journal, twoside]{ieeetran} 
\IEEEoverridecommandlockouts                              

\usepackage{graphics} 
\usepackage{amsmath} 
\usepackage{amssymb}  
\usepackage{siunitx}
\usepackage{mathtools}
\usepackage{booktabs}
\usepackage{multirow}
\usepackage[ruled]{algorithm2e}
\usepackage{subcaption} 
\usepackage{tikz}
\usepackage[bottom=57pt,top=54pt, left=54pt, right=54pt]{geometry}   

\usepackage{adjustbox}
\usepackage{pbox}
\usepackage{dblfloatfix}
\usepackage{graphicx}
\usepackage{hyperref}
\usepackage{caption}
\usepackage[nospace]{cite}

\usepackage{array,multirow,graphicx}
\usepackage{makecell}
\usepackage[T1]{fontenc}
\usepackage{microtype}
\usepackage{graphicx}
\usepackage{tikz}
\usetikzlibrary{positioning}
\usepackage{pgfplots}
\usepackage{xspace}

\newcommand{\etal}{et~al.}
\newcommand{\airsimOne}{Livingroom-Kitchen\xspace}
\newcommand{\airsimTwo}{Bedroom-Office\xspace}
\newcommand{\airsimThree}{Bathroom-Other\xspace}
\usepackage{soul}

\newcommand{\changed}[1]{\textcolor{black}{#1}}

\newenvironment{changedBlock}
    {\color{black}
    }
    { 
    \color{black}
    }

 \newcommand{\removed}[1]{}
\begin{document}

\begin{minipage}{\textwidth} \copyright 2022 IEEE. Personal use of this material is permitted. Permission from IEEE must be obtained for all other uses, in any current or future media, including reprinting/republishing this material for advertising or promotional purposes, creating new collective works, for resale or redistribution to servers or lists, or reuse of any copyrighted component of this work in other works.\\ Please cite this paper as:\\
\begin{verbatim} 
@inproceedings{
  Zurbrgg2022EmbodiedAD, 
  title = "Embodied Active Domain Adaptation for 
         Semantic Segmentation via Informative Path Planning",
  author = {Zurbrügg, René and Blum, Hermann 
          and Cadena, Cesar and Siegwart, Roland and Schmid, Lukas", 
  booktitle = "2022 IEEE Robotics and Automation Letters (RA-L)",
  year = 2022; 
} \end{verbatim} \end{minipage}

\title{
Embodied Active Domain Adaptation for Semantic Segmentation via Informative Path Planning
}

\author{René Zurbrügg$^{1}$, Hermann Blum$^{1}$, Cesar Cadena$^{1}$, Roland Siegwart$^{1}$, and Lukas Schmid$^{1}$


\thanks{This work was supported by funding from the Microsoft Swiss Joint Research Center and the HILTI Group.}%
\thanks{$^{1}$The authors are with Autonomous Systems Lab, ETH Z\"urich, Switzerland
{\tt \{zrene, blumh, cesarc, rsiegwart, schmluk\}@ethz.ch}}%
}%

\maketitle

\begin{abstract}
This work presents an embodied agent that can adapt its semantic segmentation network to new indoor environments in a fully autonomous way. 
Because semantic segmentation networks fail to generalize well to unseen environments, the agent collects images of the new environment which are then used for self-supervised domain adaptation.
We formulate this as an informative path planning problem, and present a novel information gain that leverages uncertainty extracted from the semantic model to safely collect relevant data.
As domain adaptation progresses, these uncertainties change over time and the rapid learning feedback of our system drives the agent to collect different data.
Experiments show that our method adapts to new environments faster and with higher final performance compared to an exploration objective, and can successfully be deployed to real-world environments on physical robots.

\end{abstract}


\section{Introduction}
\label{sec:introduction}
\IEEEPARstart{S}{cene} understanding is a core building block for robotic agents to interact intelligently with their environment\changed{, and is prerequisite for a number of applications ranging from service to construction robotics, or autonomous driving.} This understanding is commonly driven by deep-learnt object detection \changed{\cite{he2018mask, Gupta2014IndoorSU} and/}or semantic segmentation \changed{ \cite{refinenet,Pan2020CrossViewSS, Gupta2014IndoorSU}}. 
However, existing approaches degrade severely when evaluated on conditions that differ from those seen during training. Change in conditions can stem from varying weather conditions, changes in environments, or skill transfer from simulation to real world. These issues severely limit the application and deployment of robotic agents and scene understanding networks to new tasks and domains.

Overcoming the so called \emph{domain gap} proves to be a challenging task ~\cite{wilson2020survey}. Domain adaptation is often treated as the problem of transferring knowledge from a static source dataset to an also static target dataset, collected from the new domain. However, collecting and annotating new datasets for every domain is immensely costly.
One family of approaches tackles this issue by operating on limited or without supervision~\changed{\cite{depth_estimation_pseudo, Uncertainty_Reduction_in_SS,Yang2020FDAFD, Hoyer2021DAFormerIN}}. However, while this eases the burden of annotation, it does not address the problem of data collection and requires suitable training data for every new domain. 

Recent advances in robotics enable active interaction with the target domain, which allows for various new methods to be developed based on \emph{embodiment}, that is, leveraging active robot movement in order to collect data and learn from the environment.
A line of works~\cite{chaplot2021seal, embodied_active_learning, Chaplot2020SemanticCF} has shown that active robot motion can collect more useful data for domain adaptation tasks compared to pure exploration objectives.
These works train policies through Reinforcement Learning (RL) for both moving the agent, and deciding for which images to request annotations to re-train the segmentation module. Like every methodology, this comes with limitations.
\removed{However, the limitations of RL are amplified in the embodied domain adaptation task.}
First, the policies need to be trained in simulation, introducing a new domain shift that results in RL agents often underperforming in the real world \cite{Zhao2020SimtoRealTI}. 
Second, it is often hard to guarantee the safety and applicability of the learned policy in unknown and challenging environments.
 
To overcome these issues, we formulate active domain adaptation as an informative path planning (IPP) problem, combining the advantages of classical and learning-based methods.
Specifically, we present a system that uses uncertainty estimates of the semantic segmentation module as information measure.
This allows updating of the semantic module during the mission, as opposed to requiring to retrain the motion policy for every new semantic module.
The estimated uncertainty is integrated into a dense semantic map.
We then propose a novel information gain for sampling-based IPP, leveraging the uncertainty as the planning objective.
This guarantees that the robot stays safe in previously unknown environments, while identifying a series of view points with high model uncertainty. 
These are then utilized to train the network in a self-supervised fashion, using the spatial consistency of the map as the supervision signal.\removed{, thus not requiring any annotations for the new domain.}

Our proposed system completes the missing steps towards fully autonomous embodied domain adaption. 
Uncertainty-based planning allows for multiple perception-action-learning loops, where the planning naturally evolves with the semantic model's rising confidence. It can also revisit uncertain parts of a scene\removed{that remain uncertain after the first round of data collection}. 
The presented system can be placed in a new environment and then left alone to adapt its semantic module autonomously.
In summary, our contributions are:
\begin{enumerate}
    \item We formulate active domain adaptation as an IPP problem, and present a novel information gain for sampling-based IPP to gather data from areas of high uncertainty of a given semantic segmentation model.
    \item  We integrate this planner into a fully autonomous robotic system for active self-supervised domain adaptation of semantic segmentation and show that it operates robustly in real-time on a low-cost mobile robot.
    \item We show in thorough experimental evaluation that our system is able to autonomously improve segmentation IoU by an average of 36\% over multiple learning loops, especially increasing the performance on under-represented classes. We will make the system available as open-source \footnote{\url{https://github.com/ethz-asl/active_learning_for_segmentation}}.
\end{enumerate}


\section{Related Work}
\label{sec:rel_work}
\subsection{Online Informative Path Planning}
IPP aims to find paths in a potentially unknown environment that maximize an objective function while respecting constraints such as dynamic feasibility.
A rich variety of approaches exist, which largely can be divided into optimization \cite{optim_path_p, popovic2020informative, Bircher2016ThreedimensionalCP} or sampling \cite{zhu2021onlineIP, parikh2020rapid, sampling_based_unknown, isler2016information} based planners. 
While both approaches have been applied to online IPP in unknown environments \cite{popovic2020informative, zhu2021onlineIP, sampling_based_unknown}, sampling-based methods have the advantage that they can operate efficiently on complex 3D maps and naturally allow for arbitrary gain formulations. 
Different gains have been proposed for a variety of applications, operating on diverse map representations.
Isler~\etal~\cite{isler2016information} propose a set of information gains to speed up volumetric exploration on occupancy maps.
Bircher~\etal~\cite{Bircher2016ThreedimensionalCP} use the observable mesh area of a known surface to incrementally compute an exploration and coverage path.
Schmid~\etal~\cite{sampling_based_unknown} leverage the Truncated Signed Distance Field (TSDF) update rule to propose a gain for accurate surface reconstruction.
Parikh~\etal~\cite{parikh2020rapid} present a method for semantic mapping, using histograms of view orientations as a heuristic for semantic segmentation quality.
Popovic~\etal~\cite{Popovic2020AnIP} actively create a semantic 2D map to label crops and weeds, using a heuristic model of the uncertainty of the network. 
Other works use view planning for information gathering.
\cite{zhu2021onlineIP, popovic2020informative} use Gaussian Processes (GP) to model information for surface inspection \cite{zhu2021onlineIP} and crop monitoring \cite{popovic2020informative}, identifying paths that reduce the GP uncertainty.
However, to the best of our knowledge, no IPP approaches for embodied domain adaptation have yet been proposed.
\vspace{-4mm}
\subsection{Embodied Active Learning}
Embodied learning has been of interest to train object detection \cite{Chaplot2020SemanticCF,chaplot2021seal} or semantic segmentation networks \cite{Nilsson2021EmbodiedVA}. Note that we focus on methods aiming to train a semantic network using image supervision, which excludes end-to-end training for image- or point-goal navigation. 
\changed{Chaplot} \etal~\cite{Chaplot2020SemanticCF} train a motion policy with RL to collect images, which are then annotated and used to train an object detector. The learning objective is modelled by the inconsistency of the detector predictions.
Similarly, Nilsson~\etal~\cite{Nilsson2021EmbodiedVA}  train a RL agent to explore a 3D environment to improve a semantic segmentation network. The agent can request pseudo labels using optical flow or ground truth annotations to optimize a reward based on the improvement of the final network. 
Both methods above require human annotations.
Parallel to this work, Chaplot~\etal~\cite{chaplot2021seal} combine control and perception in order to learn an exploration policy and simultaneously improve a Mask R-CNN network. Both networks are trained in a self-supervised manner. Similar to our work and~\cite{frey2021continual}, they use a semantic map as pseudo labels. The motion policy is trained using a curiosity signal, which aims to maximize the amount of voxels with a high confidence score after complete exploration. A disadvantage of this approach is that the information gain is modelled within the RL policy \changed{which utilizes the predicted confidence of the segmentation networks and thus requires expensive re-training if the model and the respective uncertainty distribution changes}. In contrast to this, our method can operate on any given semantic model, allowing for rapid feedback of the network in continuous planning-learning cycles.
\vspace{-4mm}
\subsection{Uncertainty in Semantic Segmentation}
Uncertainty is typically divided into \emph{aleatoric} uncertainty arising from noise in the data, and \emph{epistemic} uncertainty which stems directly from the model itself. 
While e.g. GPs often come with a notion of uncertainty by design, CNNs do not directly provide reliable uncertainty information~\cite{Guo2017-kg}.
Different architectural changes, e.g. Bayesian deep learning \cite{baydeeplearning,deeplearning_bayes_ag} or deep GPs~\cite{deepgaussproc}, are currently being explored, but their scalability and quality is debated \cite{wenzel2020good}.
Extracting uncertainty estimates from latent representations \cite{postels2021hidden} has proven to be advantageous as the method can directly be utilized with several state-of-the-art CNNs and provides fast inference times, therefore making a combination with high speed CNNs feasible.
\vspace{-4mm}
\subsection{3D Semantic Mapping}
3D semantic mapping aims to simultaneously reconstruct a dense map of the environment while estimating the semantic label of each surface element. To achieve this, current solutions typically fuse semantic predictions and depth over multiple frames.
A large variety of systems exist \cite{grinvald2019volumetric, schmid2021panoptic, McCormac2017SemanticFusionD3,  Rosinol2020KimeraAO}, utilizing various methods to fuse geometric and semantic information. To fuse predictions over frames, these methods use counts or the softmax confidence to track the label distribution. We are unaware of works that explicitly capture the epistemic uncertainty of the segmentation model, which can be helpful to find informative viewpoints.

\begin{figure*}[tb]
\vspace{-5mm}
    \centering
    \includegraphics[width = 0.9\textwidth]{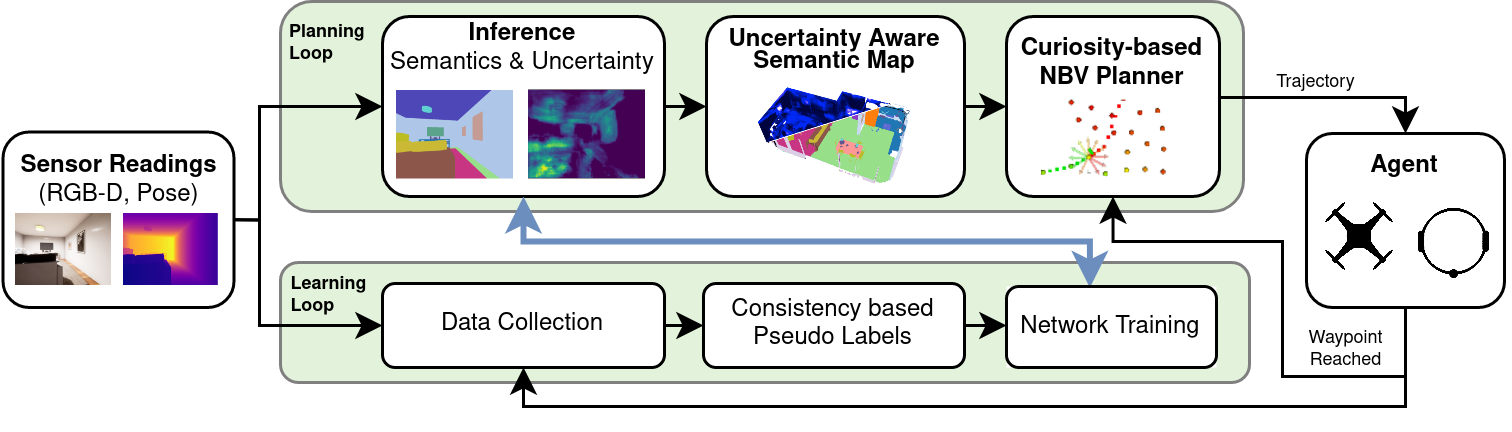}
    \caption{Overview of our embodied active learning approach. It consists of two separate loops. 
    For planning, we predict the semantic labels and the network uncertainties for each input and integrate them into the 3D map. This map is then used by the planner to move the robot and collect images of uncertain objects from different view points. 
    The learning loop captures images for training. Once a predefined amount of images have been collected, the planning loop is halted and the semantic segmentation network is re-trained. We use self supervised pseudo labels obtained by projecting the semantic map onto the training images.}
    \label{fig:overview}
\end{figure*}

\section{Approach} 
\label{sec:approach}
Our approach consists of two main loops, shown in Fig.~\ref{fig:overview}. The planning loop governs the reconstruction of the semantic map and the active movement of the robot. For each input image, we infer semantic and uncertainty information and fuse them into a 3D map. Our sampling-based path planner uses this map to select informative view points. Along the trajectory and when reaching a target, RGB-D images are captured by the learning loop. After sufficient images have been collected, we halt the planner and re-train the semantic network and its uncertainty estimator. The planning loop then resumes and the robot continues to autonomously adapt to the environment. 

\vspace{-4mm}
\subsection{Uncertainty Aware Semantic Mapping}
To estimate epistemic uncertainty of the semantic segmentation, we use a scalable method proposed by Postels et al. \cite{postels2021hidden}.
\begin{changedBlock}
Let $z_i$ denote the latent representation shortly before the logits, and $y_i, i \in [0,N]$ a possible semantic class assignment. The log likelihood of $z_i$ can be expressed as $-\log p(z_i) = -\log \sum_j p(z_i|y_j)p(y_j)$ where $p(z_j|y_i)$ is calculated using a Gaussian Mixture Model (GMM) and $p(y_i)$ is inferred based on the models output on the training data. Additionally, since latent representations $z_i$ are often very high dimensional, principal component analysis (PCA) is applied in order to reduce the dimensional complexity. 
\end{changedBlock}

Since the likelihood values typically vary for different networks and the goal is to identify views of particularly high model uncertainty, we normalize $-\log p(z_i)$ with respect to the uncertainty distribution of the current network.
To this end, we fit a Gaussian distribution \changed{ $(\mu, \sigma)$} to the log-likelihood values observed so far and compute a quantile threshold $\delta_a$, such that the probability of observing an uncertainty measurement larger than $\delta_a$ should be equal to 20\% \changed{, i.e. $\delta_a = \mu + \sigma \cdot \Phi^{-1}(0.2)$}.
The uncertainty estimate is then transformed to the normalized uncertainty $u_{pred}$:

\begin{equation}
u_{pred}(u)  = \begin{cases}
        1, & \text{if } u > u_{max}  \\
        \frac{u - \delta_a}{u_{max} - \delta_a}, & \text{if } \delta_a \leq u \leq u_{max}  \\
        0, & \text{else}
        \end{cases}
\end{equation}
where $u_{max}$ denotes the largest uncertainty value that was observed during the fitting process.

\changed{We base our mapping system on Panoptic Mapping \cite{schmid2021panoptic}, extending the single-TSDF version for semantic segmentation of \cite{schmid2021panoptic} to further integrate uncertainty estimates into the semantic map.}
Each voxel stores a counting-based distribution over class assignments, an uncertainty value $u_{map}$, and a discount value $\tau_{map}$ \changed{used for path planning (Sec.~\ref{sec:IPP})}.
To integrate new uncertainty predictions $u_{pred}(v)$ of a voxel $v$, we update current values $u_{map}^t$ using an exponential filter:

\begin{equation}
u_{map}^{t+1}(v) = \lambda \cdot u_{map}^{t}(v) + (1 - \lambda)\cdot u_{pred}(v) 
\end{equation}

The parameter $\lambda$ controls how much initially uncertain objects are explored. When overwriting voxels ($\lambda = 0$), an uncertain object will be abandoned as soon as a perspective with a confident prediction has been found. Exponential filtering with $\lambda > 0$ encourages the robot to look at an object multiple times. We empirically choose $\lambda=0.5$.
\vspace{-4mm}
\subsection{Informative Path Planning} 
\label{sec:IPP}
The goal of the planner is to find a trajectory of view points covering unseen and/or uncertain voxels. We propose a sampling-based next best view (NBV) planner, based on the implementation of \cite{sampling_based_unknown}. It continuously expands and maintains a single tree of view points with respective gain and cost values. The NBV is found by computing the global normalized value \cite{sampling_based_unknown}, selecting the sub-tree with the highest accumulated gain per cost. 
We propose to use the following generic information gain formulation. Given the set of unobserved ($\mathbb{U}$) and surface ($\mathbb{S}$) voxels, the gain $g$ for a view point $\xi$ is computed as:

\begin{equation}
g(\xi) = \sum_{v \in \textrm{Vis}(\xi)} \begin{cases}   
    \tau(v) \cdot u_{map}(v) , & \text{if } v \in \mathbb{S} \\
    \alpha_{u}, & \text{if } v \in \mathbb{U}\\
     0, & \text{else}
\end{cases}
\end{equation}

where $\textrm{Vis}(\xi)$ denotes the set of voxels visible from $\xi$ and $\tau(v)$ is the discount factor. The parameter $\alpha_u$ controls the relative importance of the uncertainty and exploration objective, where $\alpha_u\to\infty$ yields a pure exploration planner, which discards any uncertainty values and only explores unseen voxels, and $\alpha_u =0$ results in a pure uncertainty planner with no extrinsic motivation to explore.

While in a perfect online learning setup, the uncertainty of each voxel would directly be reduced by the already collected data, classical bundle-based training does not allow for immediate feedback after every frame. We therefore introduce a \emph{discount factor} $\tau(v)$, in order to approximate the expected uncertainty decrease induced by previous observations.  
We model the uncertainty decrease for an observed object as proportional to the amount of annotated pixels used for training. Since the resolution of an object in the image scales $\propto d^{-2}$ with the depth $d$, we use $\tau_{obs}(v) = \max(d,d_{min})^{-2}$ and compute the cumulative impact factor as 
\begin{align}
\tau(v) = \frac{\tau_{obs}(v)}{\tau_{map}(v) + \tau_{obs}(v)} \\
\tau_{map}^{t+1}(v) = \tau_{map}^t(v) + \tau_{obs}(v)
\end{align} 
We find that his leads to better performance than a constant observation model $\tau_{obs}(v) = 1$. We use a minimal distance $d_{min} = 1$m, to prevent moving too close to an object.
\vspace{-4mm}
\subsection{Self-supervised Network Training}
After gathering images from uncertain view points, we follow Frey~\etal~\cite{frey2021continual} to create self-supervised annotations for the collected images. The spatial consistency of the semantic map is utilized as supervision signal by rendering the highest probability class of the semantic map, yielding completely self-supervised annotations.

We train the semantic segmentation network multiple times during the exploration of the environment. This allows for better feedback between the network and the planner. After collecting a predefined number $N_B$ of images, called \emph{bundle}, the planner and data collection is paused. For each image of the bundle, we recompute the pseudo labels based on the current state of the semantic map.


\section{Experimental Setup}

\subsection{Simulation Environments}
\label{sec:simulation_setup}

\begin{figure}[bt]
    \centering
    \includegraphics[width=0.8\linewidth]{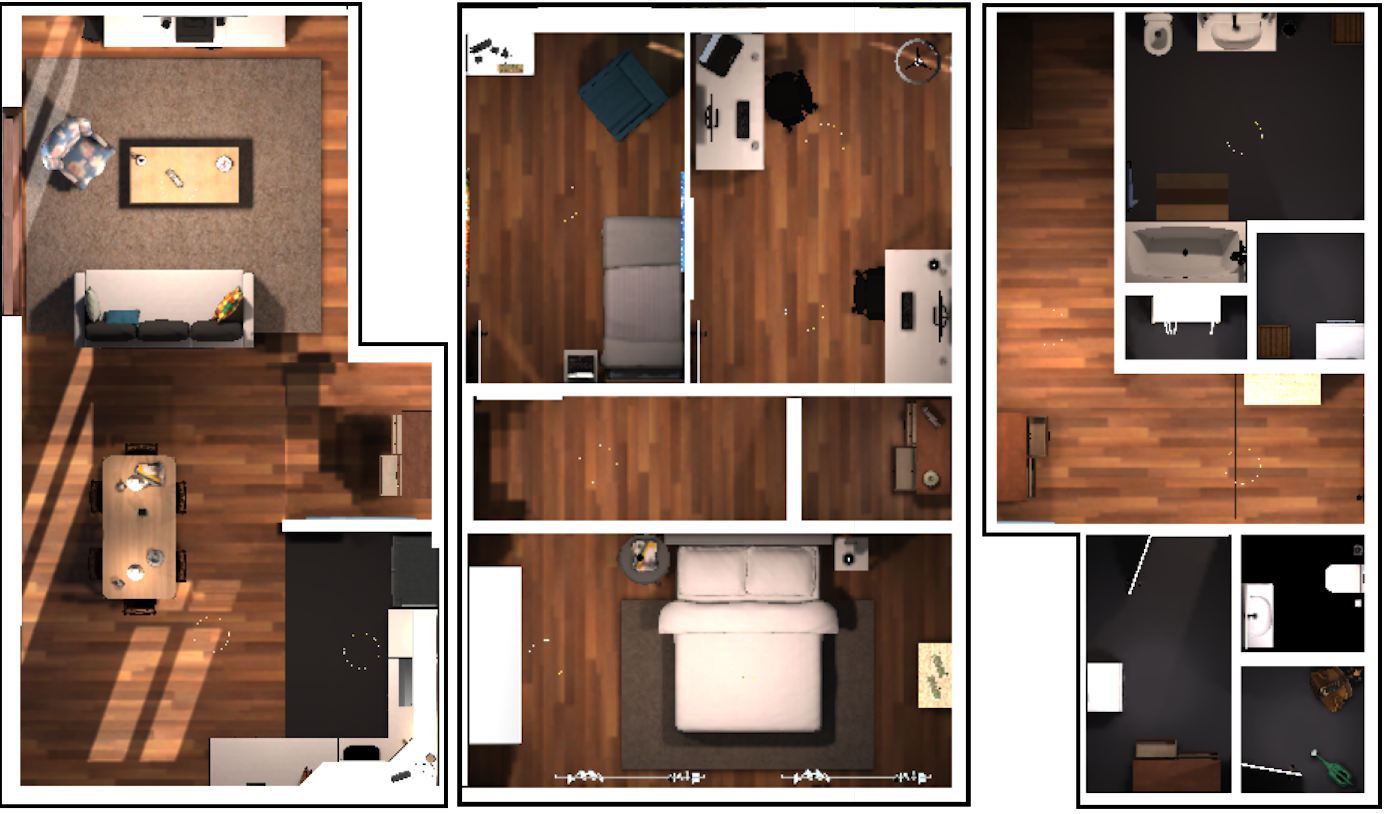}
    \caption{Overview of the three different environments \airsimOne, \airsimTwo, and \airsimThree (left to right).}
    \label{fig:overview_unreal}
\end{figure}

Since embodied domain adaptation requires interaction with the environment, a simulator rather than a dataset is required.
Perception pipelines are highly dependent on high-fidelity visual simulation and accurate ground-truth for evaluation. We use the simulation pipeline of \cite{schmid2021unified} to create photo-realistic worlds based on Unreal Engine 4\footnote{\url{https://www.unrealengine.com/en-US/}} (UE4) and AirSim \cite{airsim2017fsr}.

To evaluate our method in diverse settings, we create three different environments called \emph{\airsimOne}, \emph{\airsimTwo}, and \emph{\airsimThree}. They each contain different objects and span 47, 52, and 44m$^2$, respectively. An overview of the scenes is given in Fig.~\ref{fig:overview_unreal}.
We use $\alpha_u = 0.001$ in simulation.

We further evaluate our approach on the Matterport 3D dataset \cite{Matterport3D}, using the habitat simulator \cite{Savva2019HabitatAP}. These environments have the advantage, that the layout of each scene is captured from a real world scan. However, it also has the large disadvantage that the rendering of the scene is notably less realistic, including reconstruction artifacts and missing areas. Furthermore, the available ground truth annotation can be noisy and erroneous at times. To compensate for the correspondingly higher uncertainty values, we use $\alpha_u = 0.02$ in the Matterport environments.

We evaluate our method, both in simulation and the real world, on a ground platform. We therefore restrict the planning space to the XY-plane and yaw. Note that the exact same system can also be run in 3D, e.g. using a flying platform.

\vspace{-3mm}
\subsection{Metrics and Baselines}
For each experiment, we measure the segmentation quality as mean Intersection over Union (mIoU), evaluated over the 40 NYU\footnote{The classes Pillow/Sofa and Counter/Cabinet are merged for the \airsimOne environment, as they are part of identical object meshes.} \cite{nyu} classes as well as the eigen13 classes as defined in \cite{Couprie2013IndoorSS}. If a certain class (e.g. person) is not part of the test set, we ignore predictions for this class. Each method is evaluated on a test set of 120 uniformly sampled poses in each environment.
To account for the stochasticity in planning and network training, each method is run 7 times and the mean and standard deviation is reported.

\textcolor{black}{In the following, we refer to our method as \emph{Curiosity}. Similar to curiosity in RL, we incorporate an intrinsic gain based on model uncertainty in our path planning formulation.}
We compare against two baseline methods. All evaluated methods use identical planners but differ in the choice of information gain formulation.
\emph{Random} uses identical sampling as our method, but chooses the NBV uniformly random. While this approach does not collide, it often does not achieve full coverage of the environment during the limited mission time.
We further compare against a pure \textcolor{black}{volumetric} \emph{Exploration} approach of \cite{sampling_based_unknown}. This is identical to $\alpha_u\to\infty$, encouraging the robot to explore the whole environment. This planner always achieves full coverage, but is lacking our uncertainty-based information gain.
\vspace{-4mm}
\subsection{Learning Setup}
We employ an iteration based learning setup, where the robot iteratively collects bundles of 150 or 300 images, depending on the size of the environment. After collecting a bundle of images, all images are added to the training dataset and we train the segmentation for 15 epochs using a constant learning rate of $10^{-5}$ and $5\cdot10^{-4}$ for the en- and de-coder respectively. As in~\cite{frey2021continual}, we utilize image augmentation (random cropping, left right flipping, brightness and gamma changes) and memory replay. The replay buffer consists of images from the ScanNet \cite{Dai2017ScanNetR3} dataset and has a size of 30\% of the training set. We further prevent overfitting through early-stopping based on a validation set of held-out pseudo labels. Finally, the PCA and GMM modules used for uncertainty estimation are re-fitted using a randomly sampled subset of features that were seen during training.

All results shown for the self-supervised training use a custom variation of the Light-Weight Refinenet \cite{refinenet} which was pretrained on the ScanNet~\cite{Dai2017ScanNetR3} dataset. We modify the Light-Weight Refinenet by replacing the batch-norm layers with group-norm layers, yielding better learning performance when trained with lower batch sizes.

\section{Results}
\label{sec:results}

\begin{figure}
    \centering
    \includegraphics[width = \linewidth]{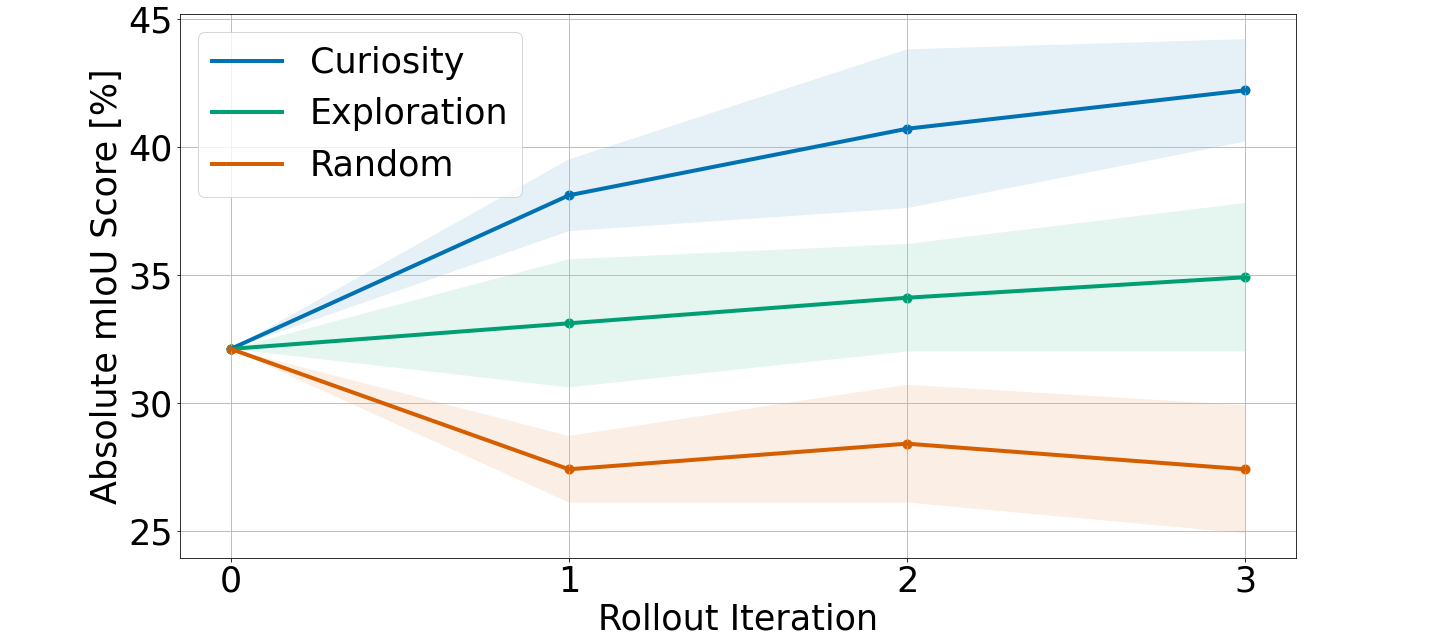}
    \caption{Model improvement after each adaptation cycle as mIoU on the NYU 40 classes. When using Exploration or Curiosity, our rapid learning feedback continuously improves the network over time. This is particularly pronounced for the Curiosity approach.}
    \label{fig:over_all_rollouts_mean_std}
\end{figure}
We thoroughly evaluate our method in various simulation and real world experiments. 
We first study the active domain adaptation performance in detail in the \airsimOne, due to its medium size and great object diversity. 
We then discuss generalization of our approach to different synthetic and reality-capture environments and robot setups.
We demonstrate the robustness of our system with experiments on an autonomous mobile robot with only limited sensing and computing capabilities.

\vspace{-4mm}
\subsection{Influence of the Trajectory on Domain Adaptation}
We compare all planners in all three UE4 environments.
Tab.~\ref{tab:miou_scores} shows the resulting mIoU scores after three consecutive learning cycles. For each cycle, 300 images are collected and training with the gathered (300, 600, and 900) images for 15 epochs is performed. For \airsimOne, which requires less exploration time, only 150 images are collected per cycle. In \airsimThree, we use $\alpha_u=0.002$ to account for the narrow room layout.

As expected, how and which data is gathered has a strong influence on the domain adaptation performance. 
Notably, not all collected data improves the model, as Random shows only little improvements and even degrades the model in certain conditions.
In line with prior work~\cite{chaplot2021seal, embodied_active_learning, Chaplot2020SemanticCF}, we find that pure exploration is a strong baseline for such data collection and significantly improves the model.
Nonetheless, we find that our curiosity information gain which trades off exploration and model uncertainty shows the best overall performance \textcolor{black}{while still yielding similar coverage and exploring the full environment}. On average, it achieves a 21.6\% stronger improvement than Exploration, leading to a 13.1 point improvement in mIoU over the pretrained model.
Due to its ability to focus on highly uncertain areas, it is the only planner that consistently improves semantic segmentation in all environments. \changed{We further provide the results for appearance and pseudo label based domain adaptation methods for the \airsimOne environment in Table \ref{tab:da_scores}.
Regardless of the chosen baseline, curios data collection achieves the highest mIoU scores. Additionally, our supervision signal outperforms all listed baselines.}

\begin{table}[tb]
    \centering
    \begin{tabular}{llcc}
    \toprule
    Environment & Method & NYU 40 & Eigen 13 \\ \midrule
    \multirow{4}{*}{\shortstack[l]{Livingroom-\\Kitchen}}    & Pretrained Model &  32.1 & 50.0 \\
         & Random &  27.4 $\pm$ 2.5 & 55.5 $\pm$ 6.8 \\
         & Exploration & 34.9 $\pm$ 2.9 & 68.5 $\pm$ 4.7 \\ 
         & Curiosity (ours) & \textbf{42.2 $\pm$ 2.0} & \textbf{69.8 $\pm$ 2.8} \\  \midrule
    \multirow{4}{*}{\shortstack[l]{Bedroom-\\Office}} & Pretrained Model &  27.7 & 40.6 \\
         & Random &  30.7 $\pm$ 3.8& 46.1 $\pm$ 9.6 \\
         & Exploration & 39.9 $\pm$ 2.2  & 59.6 $\pm$ 3.8\\
         & Curiosity (ours) & \textbf{40.6 $\pm$ 2.5}& \textbf{62.5 $\pm$ 4.6}\\ \midrule
   \multirow{4}{*}{\shortstack[l]{Bathroom-\\Other}} & Pretrained Model &  19.2 & 39.2 \\
         & Random &  20.5 $\pm$ 2.6 & 43.8 $\pm$ 1.9 \\
         & Exploration & 18.6 $\pm$  1.3 & \textbf{51.8 $\pm$ 2.0 } \\
         & Curiosity (ours) & \textbf{21.2} $\pm$ \textbf{1.9} & 50.9 $\pm$ 1.2 \\ \bottomrule
    \end{tabular}
    \caption{Final mIoU score [\%] of the network after three autonomous domain adaptation cycles. While Exploration is a strong baseline, our uncertainty-based information gain consistently further improves the model.}
    \label{tab:miou_scores}
\end{table}

\begin{table}[tb]
    \centering
    
\begin{changedBlock}
    \begin{tabular}{llcc}
    
    \toprule
    Method & Random & Exploration & Curiosity \\ \midrule
    Fourier Transfer\cite{Yang2020FDAFD} &  32.5 / 49.7 & 33.2 / 49.0 & 33.8 / 49.0   \\
    Self-training\footnotemark & \textbf{32.7} / 52.9 &  31.8 / 51.8 & 33.6 / 54.4  \\
    Uncertainty Reduction\cite{Uncertainty_Reduction_in_SS}  & \textbf{32.7} / 52.4 & 33.4 / 53.7 & 37.4 / 58.0 \\
    Spatial Consistency (ours) & 27.4 / \textbf{55.5} & \textbf{34.9} / \textbf{68.5} & \textbf{42.2 }/ \textbf{69.8}  \\
         \bottomrule
    \end{tabular}
    
\end{changedBlock}
    \caption{\changed{mIoU (NYU40 / Eigen13) [\%] on \airsimOne: Comparison of self-supervision methods using the data collected with different planners, averaged over seven runs. Most methods benefit from curios data collection.}
    }
    \label{tab:da_scores}
    \vspace{-3mm}
\end{table}

\vspace{-4mm}
\subsection{Learning Improvement Over Time}
In Fig.~\ref{fig:over_all_rollouts_mean_std} the adapted network is evaluated after each learning cycle in \airsimOne.
We observe that all methods have notable variance. 
This can be explained by multiple sources of randomness, including view point sampling, validation set selection, and model training. \textcolor{black}{We also note that due to the self supervision strategy, the final score of the model depends on the initial model performance}. 

The overall results still show significant improvements when using our method.
Most importantly, while Exploration and Random plateau after the first training cycle, our method keeps on improving the network performance.
This suggests that the uncertainty in our information gain plays a substantial role in fine-tuning a given model once the environment has been covered.
It further highlights the benefits of the proposed learning feedback, where training after every cycle and directly feeding the new model back to the planner allows our system to continuously adapt.
\footnotetext{\changed{Uses the network predictions as pseudo labels}}
\vspace{-4mm}
\subsection{Learning Improvement by Object Category}
\begin{figure}
\includegraphics[width = \linewidth]{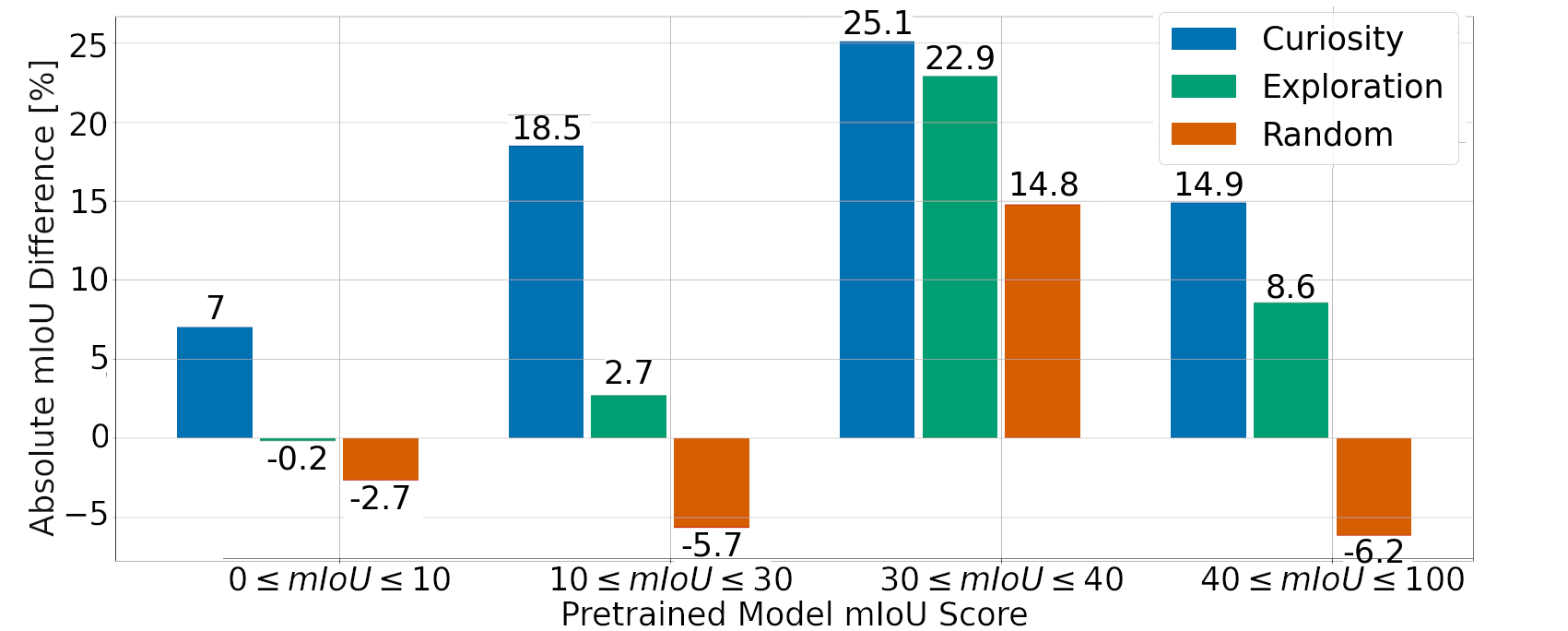}
\caption{
mIoU improvement of the final trained model grouped by the model score prior to training. Classes that the pretrained model struggles to segment (mIoU $\leq30$\%) tend to not benefit from Random or Exploration planning. In contrast, our method is able to also improve the model performance for highly uncertain classes.}    
\label{fig:miou_improve_by_grp}
\vspace{-3mm}
\end{figure}

An important feature for active domain adaptation is the ability to represent all encountered objects in the model. 
To investigate this behavior, we group the 40 NYU classes based on the segmentation performance of the pretrained network, shown in Fig.~\ref{fig:miou_improve_by_grp}.
We observe a pronounced difference between the Curiosity and Exploration planners for objects that are not well represented in the pretrained network (mIoU $\leq 30$). 
This highlights a strength of the Curiosity planner, which aims to collect images of highly uncertain objects.
These often coincide with objects the pretrained model struggles to segment correctly. 
Observing these objects from different perspectives results in a more accurate semantic reconstructions, thereby producing better pseudo labels.
This ability to also improve the network for initially uncertain classes is highly desirable and in stark contrast to the other two approaches.
This behavior could also be exploited to specifically adapt a model to certain classes of interest.

\vspace{-4mm}
\subsection{Robustness to Environment and Robot Changes}
A common risk in self-supervised domain adaptation is that models can overfit on the current environment. 
While this can be alleviated by re-adapting to new environments, the adapted network should also ensure robust autonomous operation when subject to changes. 
To measure this, the adapted networks of \airsimOne are evaluated in conditions differing from the ones during domain adaptation.
To model environmental changes, objects such as chairs, kitchen utensils, or pictures, are rearranged.
To model changes to the robot, a test set of images at varying heights is collected, yielding views which could not be reached by the ground robot during deployment. 
While for both cases the captured images still resemble the original environment, they can not directly be observed during training and model perturbations an autonomous robot might encounter. 
The resulting mIoU scores are presented in Tab.~\ref{tab:generalization_performance}.
We notice that for Exploration and Curiosity the improvements during domain adaptation translate to the modified setting.
While the mIoU improvement in the rearranged environment is reduced by 35.7\% for Exploration, Curiosity only suffers a 16.8\% improvement loss. Additionally, both planners adapt well to images sampled at different heights, most likely due to the random cropping augmentation that was employed during training.
This highlights that our learned network not only improves when evaluated on the exact same domain, but is able to deal with slight deviations.
\begin{table}
    \centering
    \begin{tabular}{lccc}
    \toprule
        Method & Adapted Domain & Rearranged & Varying Height\\ \midrule
        Pretrained & 32.1  & 36.0  & 26.7  \\  
        Random & 27.4 $\pm$ 2.5  & 30.6 $\pm$ 3.1 & 28.3 $\pm$ 2.6  \\  
        Exploration & 34.9 $\pm$ 2.9 & 37.8 $\pm$ 3.1  & 32.7 $\pm$ 3.3 \\ 
        Curiosity (ours) & \textbf{42.2 $\pm$ 2.0} &  \textbf{44.4 $\pm$ 2.7}& \textbf{39.4 $\pm$ 2.1}\\ 
        \bottomrule
    \end{tabular}
    \caption{Generalization performance as mIoU of the NYU labels. After self-supervised adaption of the network for three cycles, we evaluate it on the environment it was adapted to and two perturbed variations.
    To model environmental changes, the furniture is rearranged. To model robot changes, views at varying heights that the ground robot could not observe during training, are evaluated.}
    \label{tab:generalization_performance}
\end{table}

\vspace{-4mm}
\subsection{Evaluation in Reality-capture Simulation}
To further verify our method, we evaluate it in three environments of the Matterport 3D \cite{Matterport3D} dataset using the habitat simulator \cite{Savva2019HabitatAP}. 
We select three environments that have a complete reconstruction and are of small or medium size. The results for three runs of each method are reported in Tab.~\ref{tab:result_mp}. 
As pointed out in Sec.~\ref{sec:simulation_setup}, a disadvantage of the reality-captured scenes is the comparably poor photo-realism of rendered images, containing e.g. reconstruction artifacts and improper lighting.
This fact is reflected in the strongly reduced performance of the pretrained model.
Nonetheless, both Exploration and Curiosity show notable improvements, demonstrating that our pipeline is also able to actively adapt the model in this challenging setting. 
An advantage of Curiosity over the baseline is not evident in these scenes. 
This can likely be explained by the quality of the rendered images, which make it hard for the uncertainty estimator to produce meaningful predictions. In this case, our planner innately favors the exploration objective and achieves similar results to Exploration, not adversely affecting performance.
Another explanation is that labelling errors in the dataset potentially conceal improvements on certain objects. 

\begin{table}[]
    \centering
    \begin{tabular}{llcc}
    \toprule
Environment & Method & NYU 40 & Eigen 13 \\ \midrule
\multirow{4}{*}{17DRP5sb8fy} & Pretrained &  15.3 & 23.5\\
     & Random &  12.3$\pm$ 2.6 & 33.5 $\pm$ 8.0  \\
     & Exploration & 16.2 $\pm$ 1.9  & 39.4 $\pm$ 5.5 \\
     & Curiosity (ours) & \textbf{18.2 $\pm$ 2.5} & \textbf{44.6 $\pm$ 4.6} \\  \midrule
     \multirow{4}{*}{GdvgFV5R1Z5} & Pretrained &  15.8 & 22.6 \\
     & Random &  19.1 $\pm$ 2.2 &  27.1 $\pm$ 1.9 \\
     & Exploration & \textbf{20.8 $\pm$ 0.8} & \textbf{27.4 $\pm$ 0.8}\\
     & Curiosity (ours)& 19.5 $\pm$ 0.1 &  \textbf{27.4 $\pm$ 1.2}\\  \midrule
     \multirow{4}{*}{i5noydFURQK} & Pretrained &  20.2 & 29.7 \\
     & Random &  13.6 $\pm$ 1.9 &  33.4 $\pm$ 2.8  \\
     & Exploration & \textbf{23.5  $\pm$ 1.8 }& \textbf{42.0 $\pm$ 0.2}\\
     & Curiosity (ours) & 23.1 $\pm$ 1.0&  40.7 $\pm$ 4.5\\  \bottomrule
     
    \end{tabular}
 \caption{Results evaluated on a selection of the Matterport 3D dataset. Results denote average value over a total of three runs for each planner type. \textcolor{black}{Highest number in bold.}}
 \vspace{-4mm}
\label{tab:result_mp}
\end{table}
\vspace{-4mm}
\subsection{Real World Experiments}
We validate our approach in real world experiments on a fully autonomous mobile robot. 
We use a modified Turtlebot3\footnote{\url{https://www.turtlebot.com/}} (TB) "Burger". 
The robotic platform, shown in Fig.~\ref{fig:robot}, is equipped with a single-beam LiDAR, an Intel Realsense D435 RGB-D camera, and wheel encoders. 
We employ the TB default state estimator, GMapping, for LiDAR-based Simultaneous Localization and Mapping (SLAM).
Real-time semantic segmentation, SLAM\changed{, and network training }are executed on a portable Jetson Xavier AGX GPU.
Our uncertainty-aware 3D mapping and informative path planning are executed on a laptop Intel i7 CPU @2.60GHz.
The whole system thus runs on low-power hardware that could easily be carried by a slightly larger wheeled platform or a MAV.

\begin{figure*}[t!]
\vspace{-3mm}
\centering
\def\colwidth{0.12\textwidth}
\def\doublecolwidth{0.22\textwidth}
\def\imgwidth{0.12\textwidth}
\newcolumntype{M}[1]{>{\centering\arraybackslash}m{#1}}
\addtolength{\tabcolsep}{-4pt}
\renewcommand{\arraystretch}{0.5}

\begin{tabular}{M{\colwidth} M{\colwidth} M{\colwidth} M{\colwidth} M{\colwidth} M{\colwidth} M{\colwidth} M{\colwidth}}
 \multicolumn{8}{c}{\includegraphics[width=0.95\textwidth]{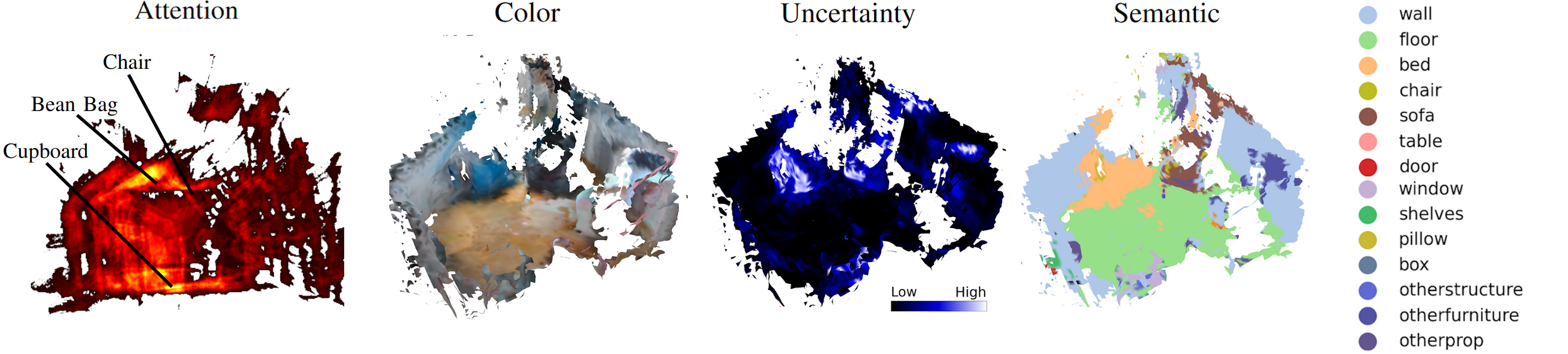}}\tabularnewline\vspace{-2mm}
  RGB image & Pretrained &  Pseudo Labels & Finetuned & RGB image & Pretrained  &  Pseudo Labels & Finetuned \tabularnewline
 \includegraphics[width=\imgwidth]{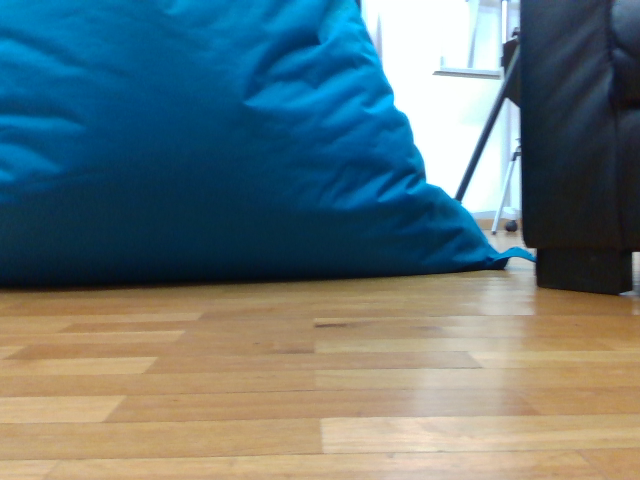} & 
 \includegraphics[width=\imgwidth]{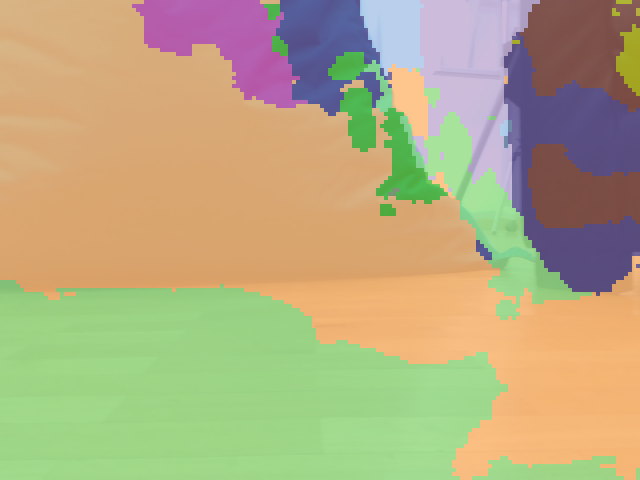} & 
 \includegraphics[width=\imgwidth]{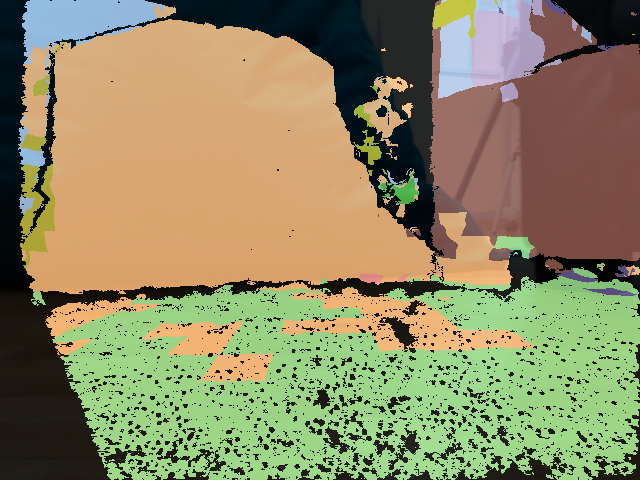} & 
 \includegraphics[width=\imgwidth]{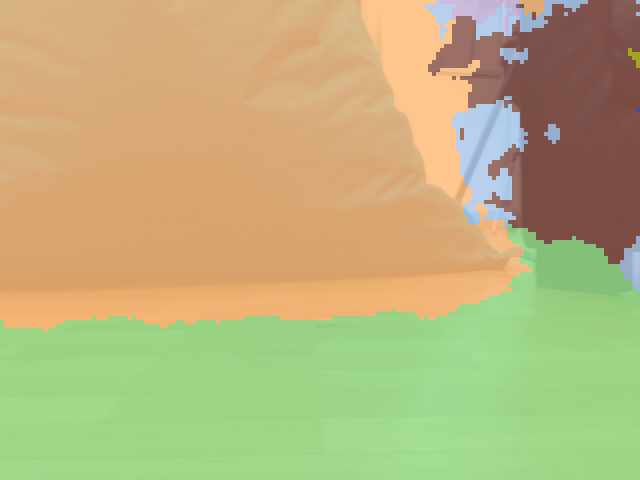} & 
 \includegraphics[width=\imgwidth]{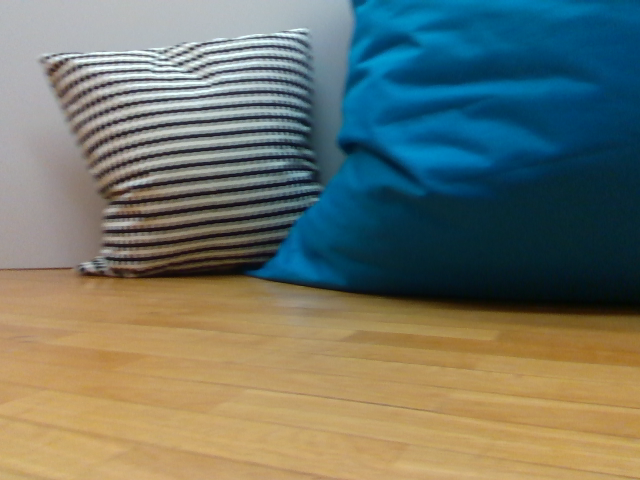} & 
 \includegraphics[width=\imgwidth]{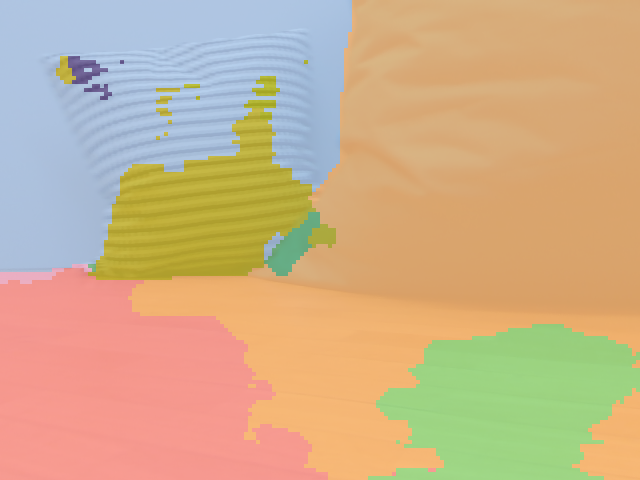} & 
 \includegraphics[width=\imgwidth]{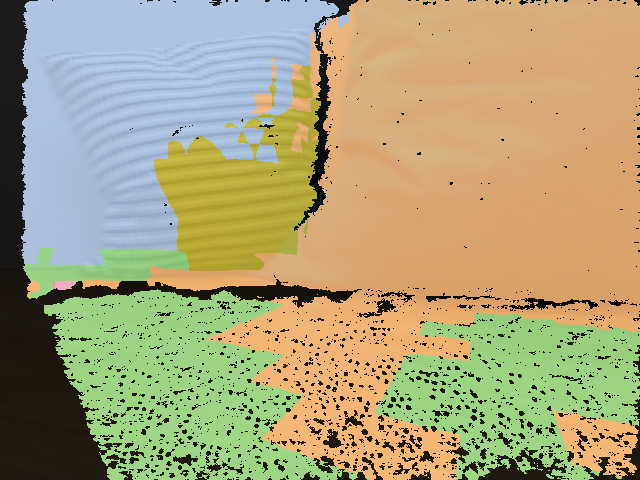} & 
 \includegraphics[width=\imgwidth]{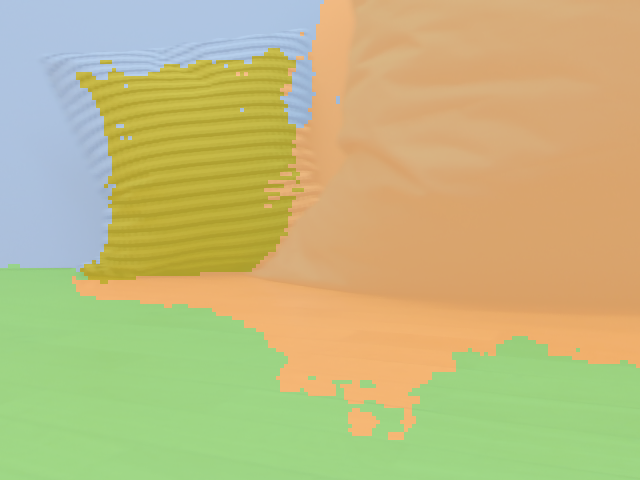}\tabularnewline
 \includegraphics[width=\imgwidth]{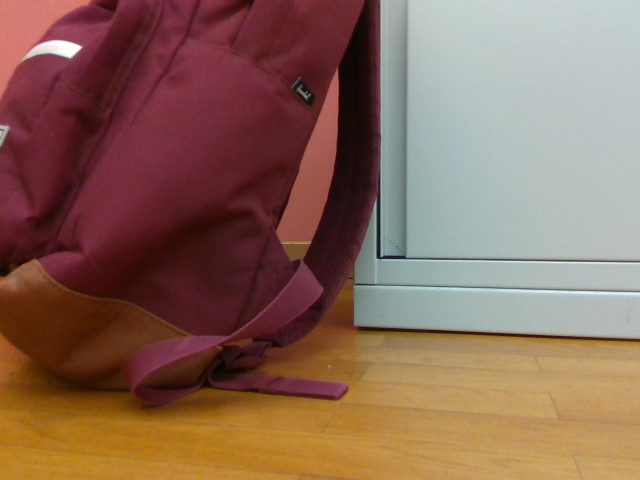} & 
 \includegraphics[width=\imgwidth]{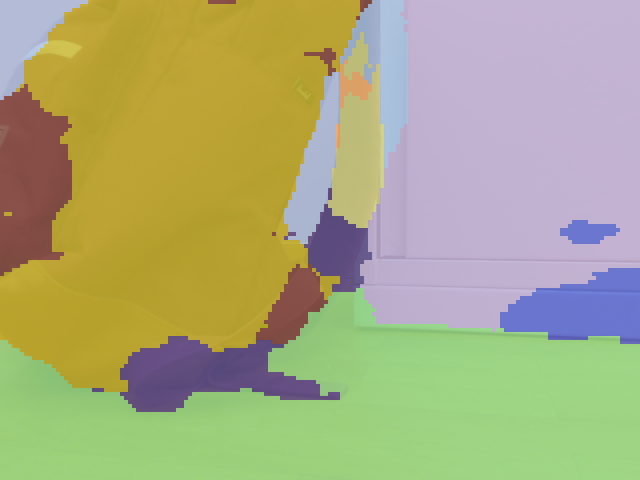} & 
 \includegraphics[width=\imgwidth]{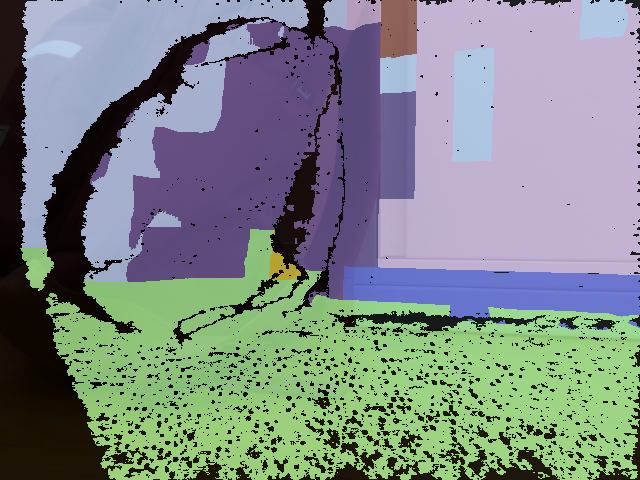} & 
 \includegraphics[width=\imgwidth]{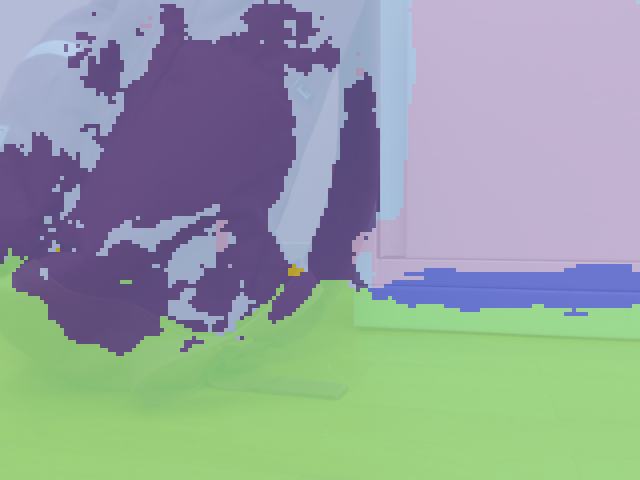} &
 
  \includegraphics[width=\imgwidth]{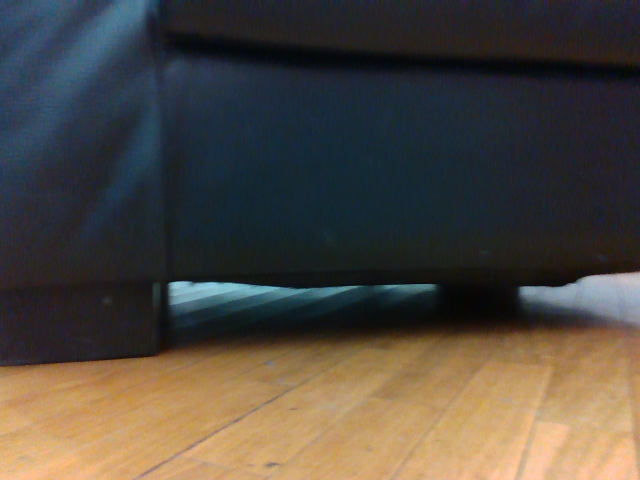} & 
 \includegraphics[width=\imgwidth]{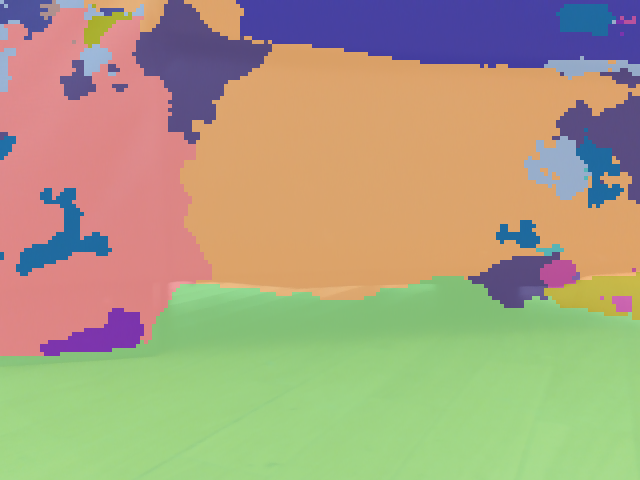} & 
 \includegraphics[width=\imgwidth]{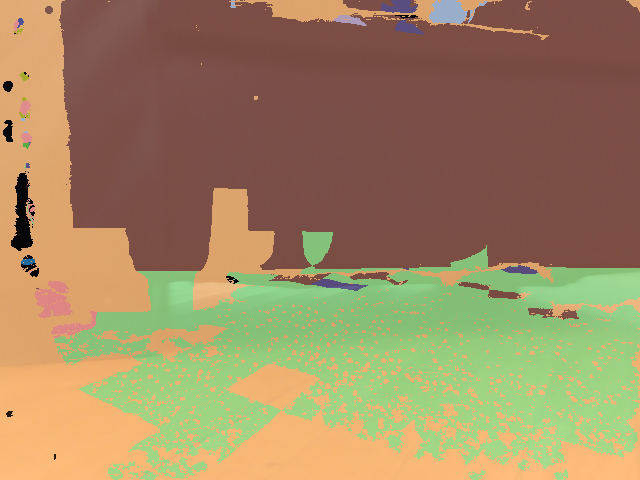} & 
 \includegraphics[width=\imgwidth]{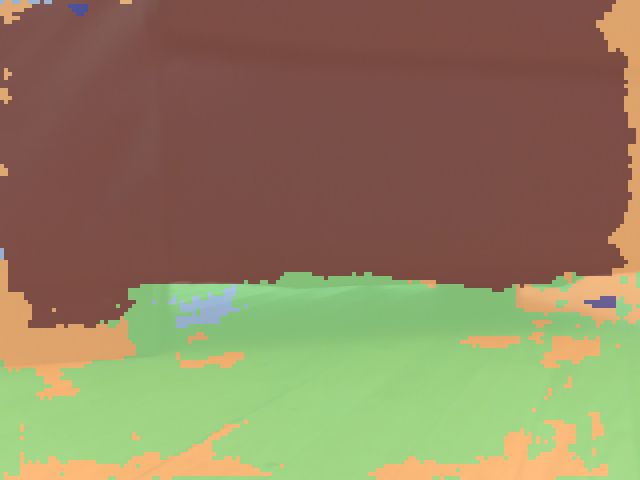} \tabularnewline
 
  \includegraphics[width=\imgwidth]{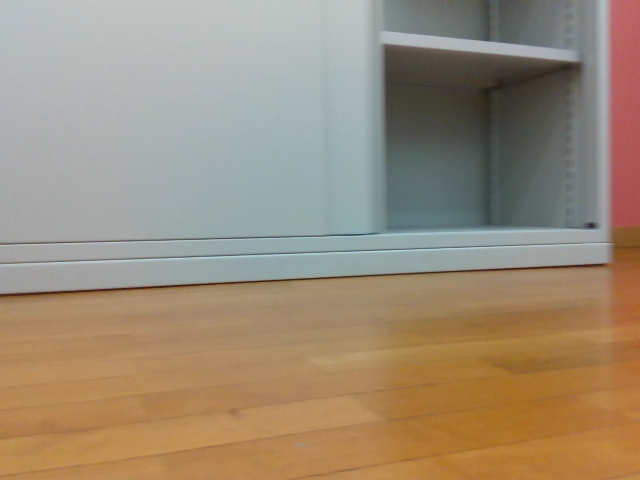} & 
 \includegraphics[width=\imgwidth]{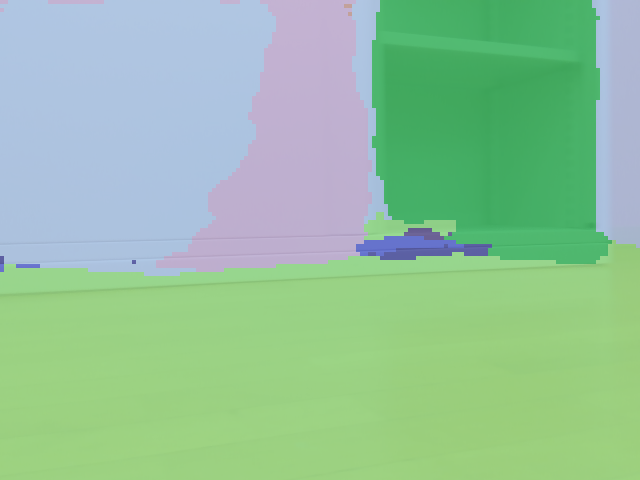} & 
 \includegraphics[width=\imgwidth]{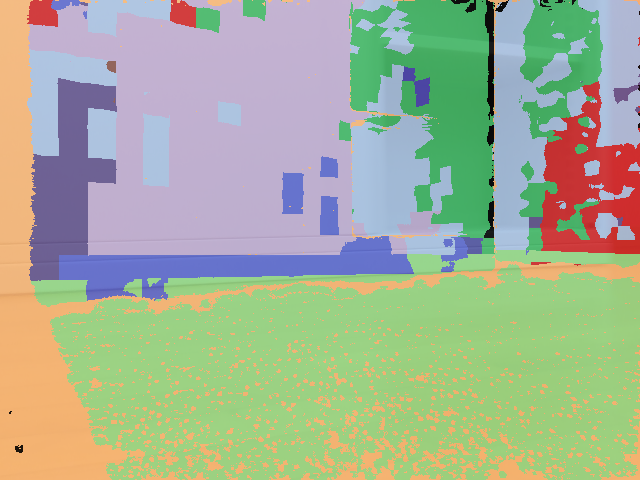} & 
 \includegraphics[width=\imgwidth]{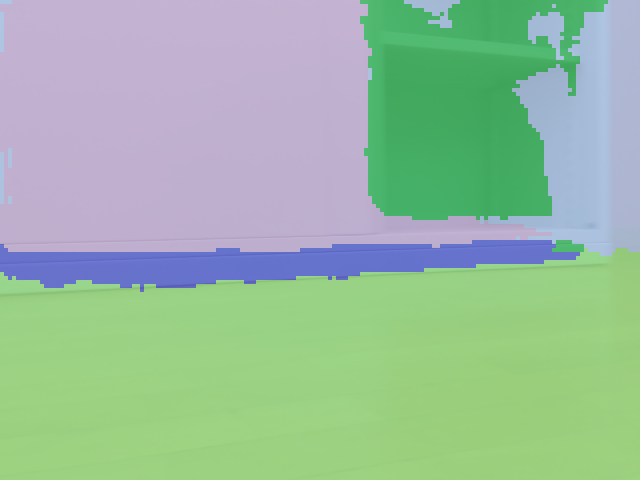} &

 \includegraphics[width=\imgwidth]{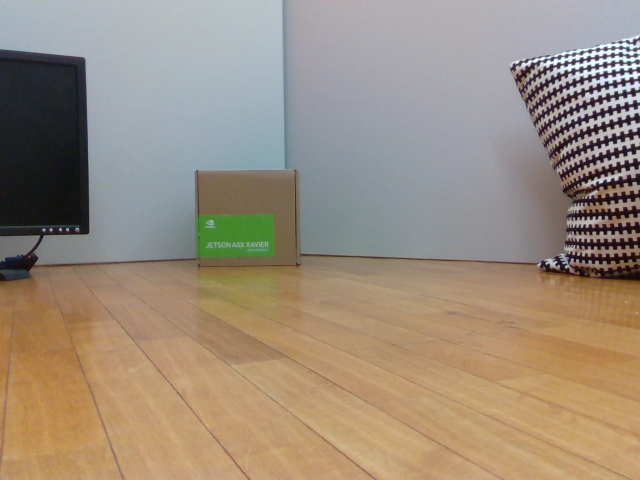} & 
 \includegraphics[width=\imgwidth]{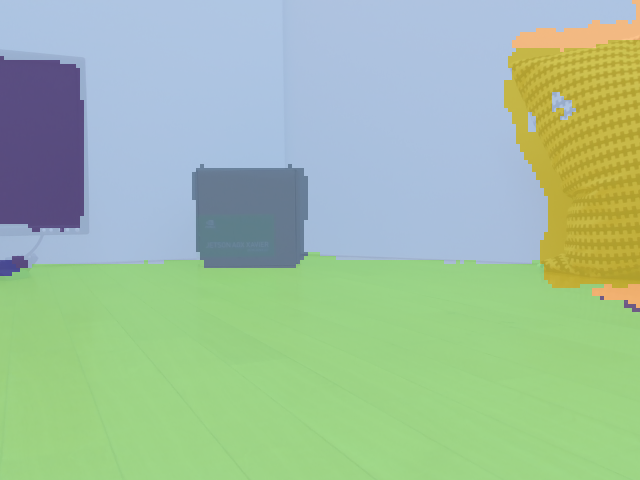} & 
 \includegraphics[width=\imgwidth]{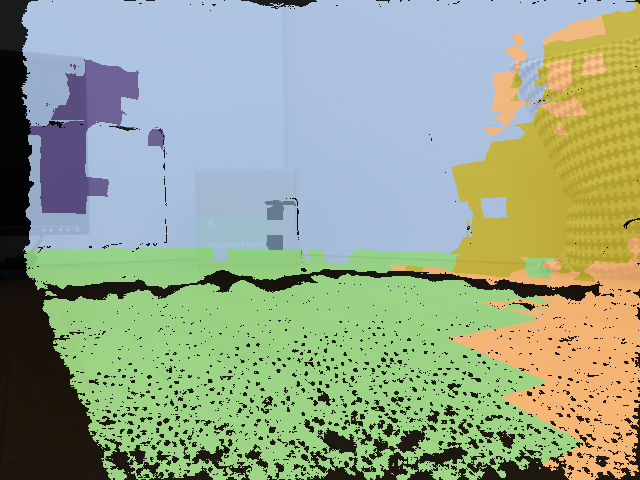} & 
 \includegraphics[width=\imgwidth]{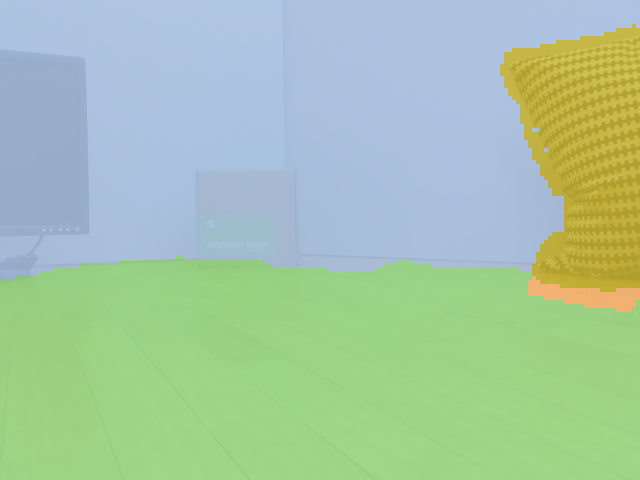}
\end{tabular}
\caption{
Top row: Overview of the attention map and reconstructed meshes. In the attentation map, points that occur in multiple views are colored brighter. Points with a height $z\leq 2$cm are omitted for clarity. Uncertain objects such as the Pillow, Bean Bag, and Cupboard are scrutinized by our planner. The multiple surfaces, e.g. on the left, show the extent of noise in the state estimation. \\Bottom area: Selected images from the real-world experiment. 
Initially highly uncertain objects such as the Bean Bag (top row), indicated by noisy predictions, are thoroughly mapped by our active system and well represented after domain adaptation. 
Miss-classified objects (center row), e.g. the backpack labeled chair or chair labeled bed, are corrected over multiple observations and correctly recognized afterwards.
Limitations of our method (bottom row) include a shelf, initially partly mislabeled as wall/window that is now only window, and small objects like the box not being captured in the map due to pose noise, which are then forgotten by the model.
We observe a general trend of our method to improve the accuracy of object boundaries during autonomous domain adaptation.}
\label{fig:real_world_images_overview}
\vspace{-3mm}
\end{figure*}

To compensate for the high sensor and localization noise of the low-cost hardware, two minor modifications are employed. 
First, only images \changed{sampled at 0.3Hz and when the robot reaches a goal pose,} rather than the whole data stream, are integrated into the semantic map. We found that these images tend to have notably better pose estimates compared to when the robot is in rapid motion \changed{such as turning}.
Second, pseudo labels are rendered by looking up the depth image in the map instead of raycasting. This way, inaccurate poses result in queries of free space, whose labels are ignored (see Fig.~\ref{fig:real_world_images_overview}), where raycasting would introduce many erroneous pseudo labels.
We deploy the robot in the indoor scene shown in Fig.~\ref{fig:real_environment} and autonomously collect 120 images using the proposed Curiosity planner. \changed{We then train the network for twelve epochs, requiring roughly 20 minutes.}
Since no groundtruth is available, extensive qualitative evaluations are reported.
The top row of Fig.~\ref{fig:real_world_images_overview} shows the final reconstructed map in color, semantics, and uncertainty, showing that our pipeline can produce meaningful maps in the real world on low-cost hardware. 

To investigate the planner behavior, Fig.~\ref{fig:real_world_images_overview} shows a heat map of the collected observations. Since the network shows high uncertainty about the Bean Bag, Chair, and Cupboard, indicated by high uncertainty values in the map of Fig.~\ref{fig:real_world_images_overview}, these objects are observed more frequently by our planner.

The resulting learning improvement is discussed in Fig.~\ref{fig:real_world_images_overview}, bottom. 
We observe that the Bean Bag and Chair were indeed highly uncertain, indicated by the noisy segmentation of these objects by the pretrained model.
After collecting data and building a good map containing these objects, the segmentation performance notably increases.
\removed{Due to the high localization and depth errors, small objects such as the cardboard box are not captured in the map and consequently forgotten by the adapted model, showing a potential limitation of the approach.}

\begin{figure}
\centering
\begin{subfigure}{.52\linewidth}
\centering
\begin{tikzpicture}
\footnotesize
\node at (0, 0) {\includegraphics[width=.4\linewidth]{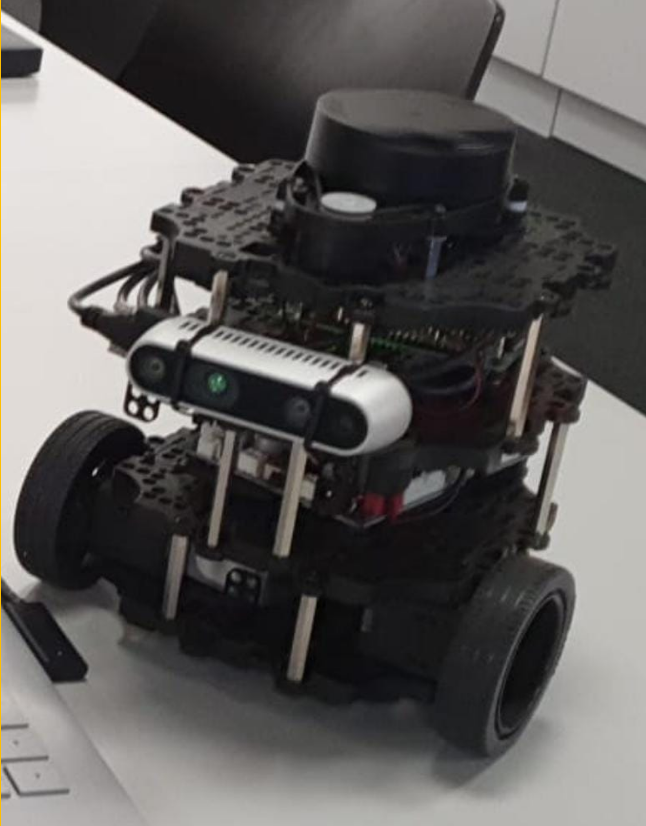}};
\node[anchor=east] (lidar) at (-1.5, 1.1) {LiDAR};
\draw[red, thick] (lidar) -- (0.0, 0.9);
\node[anchor=east] (cam) at (-1.45, .1)  {\parbox{\widthof{RGB-D}}{RGB-D\\Camera}};
\draw[red, thick] (cam) -- (-0.3, .1);

\node[anchor=east] (wheel) at (-1.6, -.8)  {\parbox{\widthof{wheel}}{Wheel\\Encoders}};
\draw[red, thick] (wheel) -- (-1, -.3);
\draw[red, thick] (wheel) -- (.3, -0.8);
    \end{tikzpicture}
    \caption{Wheeled Robotic Platform}
    \label{fig:robot}
\end{subfigure}%
\begin{subfigure}{.45\linewidth}
\includegraphics[width=.80\linewidth]{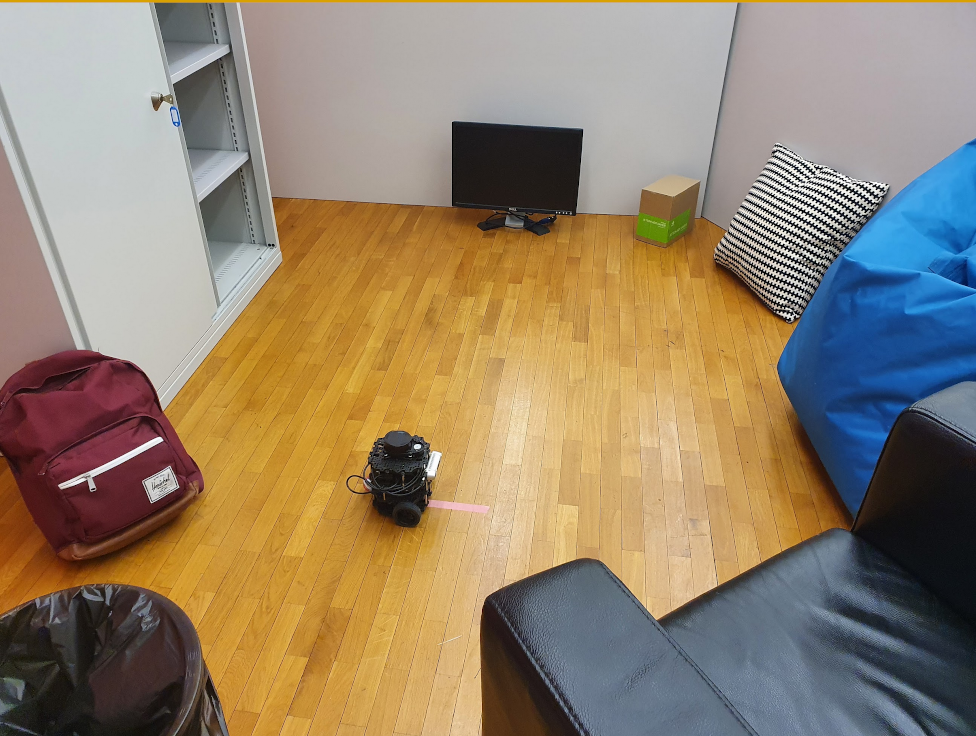}
\caption{Evaluation Environment}
\label{fig:real_environment}
\end{subfigure}%
\caption{Real world turtle bot setup.}
\label{fig:experimental_setup}
\vspace{-4mm}
\end{figure}
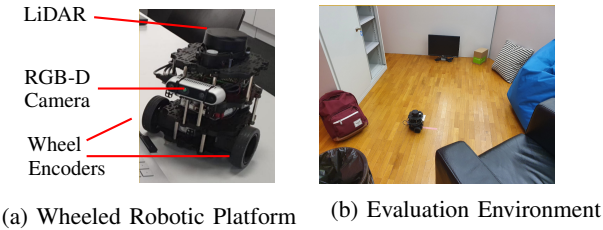
\begin{changedBlock}
\vspace{-4mm}
\subsection{Real World Limitations}
While showing good results, our method relies on an accurate reconstruction of the environment. Erroneous reconstructions arising from geometrical errors or semantic misclassifications may lead to wrong pseudo labels. This effect can be observed in the last row of Fig.\ref{fig:real_world_images_overview} where small objects such as the cardboard box are not captured in the map and consequently forgotten by the adapted model. Better depth sensors already exist and would improve reconstruction. 
Progress in uncertainty estimation could further help to filter misclassifications in the map.
\end{changedBlock}%
\section{Conclusion}
\label{sec:conclusion}
We present an embodied agent that can adapt its semantic segmentation module to unknown indoor environments in a fully autonomous fashion. We propose an IPP-based approach, that leverages epistemic uncertainty estimates integrated into a semantic map and a novel information gain formulation to safely and autonomously collect data for self-supervised training.
Our experiments show that our method outperforms pure exploration objectives with especially high margins on underrepresented classes. Fast information feedback allows our system to iteratively improve the network over time. Experiments on the Matterport3D validate \textcolor{black}{improved} adaptation on data captured from real environments, although the reduced data quality limits further quantitative evaluations.
Finally, we employ embodied domain adaptation on a physical robot with limited hardware and show that our method successfully translates to the real world.

{\small
\bibliographystyle{IEEEtran}
\bibliography{IEEEfull,root}

\begin{thebibliography}{10}
\providecommand{\url}[1]{#1}
\csname url@rmstyle\endcsname
\providecommand{\newblock}{\relax}
\providecommand{\bibinfo}[2]{#2}
\providecommand\BIBentrySTDinterwordspacing{\spaceskip=0pt\relax}
\providecommand\BIBentryALTinterwordstretchfactor{4}
\providecommand\BIBentryALTinterwordspacing{\spaceskip=\fontdimen2\font plus
\BIBentryALTinterwordstretchfactor\fontdimen3\font minus
  \fontdimen4\font\relax}
\providecommand\BIBforeignlanguage[2]{{%
\expandafter\ifx\csname l@#1\endcsname\relax
\typeout{** WARNING: IEEEtran.bst: No hyphenation pattern has been}%
\typeout{** loaded for the language `#1'. Using the pattern for}%
\typeout{** the default language instead.}%
\else
\language=\csname l@#1\endcsname
\fi
#2}}

\bibitem{he2018mask}
K.~He, G.~Gkioxari, P.~Doll{\'a}r, and R.~Girshick, ``Mask r-cnn,'' \emph{2017
  IEEE Int. Conf. on Computer Vision (ICCV)}, pp. 2980--2988, 2017.

\bibitem{Gupta2014IndoorSU}
S.~Gupta, P.~Arbel{\'a}ez, R.~B. Girshick, and J.~Malik, ``Indoor scene
  understanding with rgb-d images: Bottom-up segmentation, object detection and
  semantic segmentation,'' \emph{Int. Journal of Computer Vision}, vol. 112,
  pp. 133--149, 2014.

\bibitem{refinenet}
G.~Lin, A.~Milan, C.~Shen, and I.~D. Reid, ``Refinenet: Multi-path refinement
  networks for high-resolution semantic segmentation,'' \emph{CoRR}, vol.
  abs/1611.06612, 2016.

\bibitem{Pan2020CrossViewSS}
B.~Pan, J.~Sun, H.~Y.~T. Leung, A.~Andonian, and B.~Zhou, ``Cross-view semantic
  segmentation for sensing surroundings,'' \emph{IEEE Robotics and Automation
  Letters}, vol.~5, pp. 4867--4873, 2020.

\bibitem{wilson2020survey}
G.~Wilson and D.~J. Cook, ``A survey of unsupervised deep domain adaptation,''
  \emph{ACM Transactions on Intelligent Systems and Technology (TIST)},
  vol.~11, no.~5, pp. 1--46, 2020.

\bibitem{depth_estimation_pseudo}
Q.~Wang, D.~Dai, L.~Hoyer, O.~Fink, and L.~V. Gool, ``Domain adaptive semantic
  segmentation with self-supervised depth estimation,'' \emph{CoRR}, vol.
  abs/2104.13613, 2021.

\bibitem{Uncertainty_Reduction_in_SS}
P.~T. S and F.~Fleuret, ``Uncertainty reduction for model adaptation in
  semantic segmentation,'' in \emph{Proc.~of IEEE Conf.~on Computer Vision and
  Pattern Recognition}, June 2021, pp. 9613--9623.

\bibitem{Yang2020FDAFD}
Y.~Yang and S.~Soatto, ``Fda: Fourier domain adaptation for semantic
  segmentation,'' \emph{CVPR}, pp. 4084--4094, 2020.

\bibitem{Hoyer2021DAFormerIN}
L.~Hoyer, D.~Dai, and L.~V. Gool, ``Daformer: Improving network architectures
  and training strategies for domain-adaptive semantic segmentation,''
  \emph{ArXiv}, vol. abs/2111.14887, 2021.

\bibitem{chaplot2021seal}
D.~S. Chaplot, M.~Dalal, S.~Gupta, J.~Malik, and R.~Salakhutdinov, ``{SEAL}:
  Self-supervised embodied active learning using exploration and 3d
  consistency,'' in \emph{NeurIPS}, 2021.

\bibitem{embodied_active_learning}
D.~Nilsson, A.~Pirinen, E.~G{\"{a}}rtner, and C.~Sminchisescu, ``Embodied
  visual active learning for semantic segmentation,'' \emph{CoRR}, vol.
  abs/2012.09503, 2020.

\bibitem{Chaplot2020SemanticCF}
D.~S. Chaplot, H.~Jiang, S.~Gupta, and A.~Gupta, ``Semantic curiosity for
  active visual learning,'' in \emph{ECCV}, 2020.

\bibitem{Zhao2020SimtoRealTI}
W.~Zhao, J.~P. Queralta, and T.~Westerlund, ``Sim-to-real transfer in deep
  reinforcement learning for robotics: a survey,'' \emph{2020 IEEE Symposium
  Series on Computational Intelligence (SSCI)}, pp. 737--744, 2020.

\bibitem{optim_path_p}
K.~Alexis, C.~Papachristos, R.~Siegwart, and A.~Tzes, ``Uniform coverage
  structural inspection path–planning for micro aerial vehicles,'' in
  \emph{2015 IEEE Int.~Conf.~ on Intelligent Control (ISIC)}, 2015, pp. 59--64.

\bibitem{popovic2020informative}
M.~Popovic, T.~Vidal-Calleja, G.~Hitz, J.~Chung, I.~Sa, R.~Siegwart, and
  J.~Nieto, ``An informative path planning framework for uav-based terrain
  monitoring,'' \emph{Autonomous Robots}, vol.~44, 07 2020.

\bibitem{Bircher2016ThreedimensionalCP}
A.~Bircher, M.~Kamel, K.~Alexis, M.~Burri, P.~Oettershagen, S.~Omari,
  T.~Mantel, and R.~Y. Siegwart, ``Three-dimensional coverage path planning via
  viewpoint resampling and tour optimization for aerial robots,''
  \emph{Autonomous Robots}, vol.~40, pp. 1059--1078, 2016.

\bibitem{zhu2021onlineIP}
H.~Zhu, J.~J. Chung, N.~R.~J. Lawrance, R.~Y. Siegwart, and J.~Alonso-Mora,
  ``Online informative path planning for active information gathering of a 3d
  surface,'' \emph{IEEE Int.~Conf.~on Robotics \& Automation}, pp. 1488--1494,
  2021.

\bibitem{parikh2020rapid}
A.~Parikh, M.~W. Koch, T.~J. Blada, and S.~P. Buerger, ``Rapid autonomous
  semantic mapping,'' in \emph{IROS}, 2020.

\bibitem{sampling_based_unknown}
L.~Schmid, M.~Pantic, R.~Khanna, L.~Ott, R.~Siegwart, and J.~Nieto, ``An
  efficient sampling-based method for online informative path planning in
  unknown environments,'' \emph{IEEE Robotics and Automation Letters}, vol.~5,
  no.~2, p. 1500–1507, Apr 2020.

\bibitem{isler2016information}
S.~Isler, R.~Sabzevari, J.~Delmerico, and D.~Scaramuzza, ``An information gain
  formulation for active volumetric 3d reconstruction,'' in \emph{IEEE
  Int.~Conf.~on Robotics \& Automation}, 2016.

\bibitem{Popovic2020AnIP}
M.~Popovic, T.~A. Vidal-Calleja, G.~Hitz, J.~J. Chung, I.~Sa, R.~Y. Siegwart,
  and J.~I. Nieto, ``An informative path planning framework for uav-based
  terrain monitoring,'' \emph{ArXiv}, vol. abs/1809.03870, 2020.

\bibitem{Nilsson2021EmbodiedVA}
D.~Nilsson, A.~Pirinen, E.~G{\"a}rtner, and C.~Sminchisescu, ``Embodied visual
  active learning for semantic segmentation,'' in \emph{AAAI}, 2021.

\bibitem{frey2021continual}
J.~Frey, H.~Blum, F.~Milano, R.~Y. Siegwart, and C.~Cadena, ``Continual
  learning of semantic segmentation using complementary 2d-3d data
  representations,'' \emph{ArXiv}, vol. abs/2111.02156, 2021.

\bibitem{Guo2017-kg}
C.~Guo, G.~Pleiss, Y.~Sun, and K.~Q. Weinberger, ``On calibration of modern
  neural networks,'' in \emph{Proc. of the 34th Int. Conf. on Machine
  Learning}, ser. Proceedings of Machine Learning Research, D.~Precup and Y.~W.
  Teh, Eds.\hskip 1em plus 0.5em minus 0.4em\relax PMLR, 2017.

\bibitem{baydeeplearning}
A.~Wilson and P.~Izmailov, ``Bayesian deep learning and a probabilistic
  perspective of generalization,'' \emph{CoRR}, vol. abs/2002.08791, 2020.

\bibitem{deeplearning_bayes_ag}
K.~Wang, P.~Vicol, J.~Lucas, L.~Gu, R.~B. Grosse, and R.~S. Zemel,
  ``Adversarial distillation of bayesian neural network posteriors,''
  \emph{CoRR}, vol. abs/1806.10317, 2018.

\bibitem{deepgaussproc}
K.~Jakkala, ``Deep gaussian processes: {A} survey,'' \emph{CoRR}, vol.
  abs/2106.12135, 2021.

\bibitem{wenzel2020good}
F.~Wenzel, K.~Roth, B.~Veeling, J.~Swiatkowski, L.~Tran, S.~Mandt, J.~Snoek,
  T.~Salimans, R.~Jenatton, and S.~Nowozin, ``How good is the bayes posterior
  in deep neural networks really?'' in \emph{ICML}, 2020.

\bibitem{postels2021hidden}
J.~Postels, H.~Blum, C.~Cadena, R.~Y. Siegwart, L.~V. Gool, and F.~Tombari,
  ``Quantifying aleatoric and epistemic uncertainty using density estimation in
  latent space,'' \emph{ArXiv}, vol. abs/2012.03082, 2020.

\bibitem{grinvald2019volumetric}
M.~{Grinvald}, F.~{Furrer}, T.~{Novkovic}, J.~J. {Chung}, C.~{Cadena},
  R.~{Siegwart}, and J.~{Nieto}, ``{Volumetric Instance-Aware Semantic Mapping
  and 3D Object Discovery},'' \emph{IEEE Robotics and Automation Letters},
  2019.

\bibitem{schmid2021panoptic}
L.~Schmid, J.~Delmerico, J.~Sch{\"o}nberger, J.~Nieto, M.~Pollefeys,
  R.~Siegwart, and C.~Cadena, ``Panoptic multi-tsdfs: a flexible representation
  for online multi-resolution volumetric mapping and long-term dynamic scene
  consistency,'' \emph{ArXiv}, vol. abs/2109.10165, 2021.

\bibitem{McCormac2017SemanticFusionD3}
J.~McCormac, A.~Handa, A.~J. Davison, and S.~Leutenegger, ``Semanticfusion:
  Dense 3d semantic mapping with convolutional neural networks,'' \emph{IEEE
  Int.~Conf.~on Robotics \& Automation}, 2017.

\bibitem{Rosinol2020KimeraAO}
A.~Rosinol, M.~Abate, Y.~Chang, and L.~Carlone, ``Kimera: an open-source
  library for real-time metric-semantic localization and mapping,''
  \emph{ICRA}, 2020.

\bibitem{schmid2021unified}
L.~Schmid, V.~Reijgwart, L.~Ott, J.~Nieto, R.~Siegwart, and C.~Cadena, ``A
  unified approach for autonomous volumetric exploration of large scale
  environments under severe odometry drift,'' \emph{IEEE Robotics and
  Automation Letters}, vol.~6, no.~3, pp. 4504--4511, 2021.

\bibitem{airsim2017fsr}
S.~Shah, D.~Dey, C.~Lovett, and A.~Kapoor, ``Airsim: High-fidelity visual and
  physical simulation for autonomous vehicles,'' in \emph{Field and Service
  Robotics}, 2017.

\bibitem{Matterport3D}
A.~Chang, A.~Dai, T.~Funkhouser, M.~Halber, M.~Niessner, M.~Savva, S.~Song,
  A.~Zeng, and Y.~Zhang, ``Matterport3d: Learning from rgb-d data in indoor
  environments,'' \emph{3DV}, 2017.

\bibitem{Savva2019HabitatAP}
M.~Savva, A.~Kadian, O.~Maksymets, Y.~Zhao, E.~Wijmans, B.~Jain, J.~Straub,
  J.~Liu, V.~Koltun, J.~Malik, D.~Parikh, and D.~Batra, ``Habitat: A platform
  for embodied ai research,'' \emph{ICCV}, 2019.

\bibitem{nyu}
P.~K. Nathan~Silberman, Derek~Hoiem and R.~Fergus, ``Indoor segmentation and
  support inference from rgbd images,'' in \emph{ECCV}, 2012.

\bibitem{Couprie2013IndoorSS}
C.~Couprie, C.~Farabet, L.~Najman, and Y.~LeCun, ``Indoor semantic segmentation
  using depth information,'' \emph{CoRR}, vol. abs/1301.3572, 2013.

\bibitem{Dai2017ScanNetR3}
A.~Dai, A.~X. Chang, M.~Savva, M.~Halber, T.~A. Funkhouser, and M.~Nie{\ss}ner,
  ``Scannet: Richly-annotated 3d reconstructions of indoor scenes,''
  \emph{Proc.~of IEEE Conf.~on Computer Vision and Pattern Recognition}, pp.
  2432--2443, 2017.

\end{thebibliography}
}

\clearpage

 \end{document}